\documentclass[a4paper]{article}

\usepackage{graphicx}
\usepackage{onecolceurws}
\usepackage{amsmath}

\usepackage{times}

\usepackage{latexsym}
\usepackage{multirow}
\usepackage{longtable}

\title{Visual-Semantic Embedding Model Informed by Structured Knowledge}

\author{
Mirantha Jayathilaka \\ Department of Computer Science\\
                The University of Manchester \\ mirantha.jayathilaka@manchester.ac.uk
\and
Tingting Mu \\ Department of Computer Science\\
                The University of Manchester \\ tingting.mu@manchester.ac.uk
\and
Uli Sattler \\ Department of Computer Science\\
                The University of Manchester \\  uli.sattler@manchester.ac.uk
}

\institution{}

\begin{document}
\maketitle

\begin{abstract}

We propose a novel approach to improve a visual-semantic embedding model by incorporating concept representations captured from an external structured knowledge base. We investigate its performance on image classification under both standard and zero-shot settings. We propose two novel evaluation frameworks to analyse classification errors with respect to the class hierarchy indicated by the knowledge base. The approach is tested using the ILSVRC 2012 image dataset and a WordNet knowledge base. With respect to both standard and zero-shot image classification, our approach shows superior performance compared with the original approach, which uses word embeddings. 
 
\end{abstract}

\section{Introduction}

Current state-of-the-art deep learning approaches \cite{krizhevsky2012imagenet} in computer vision are often criticised for their inability to obtain a generalised image understanding \cite{marcus2018deep} and interpretability of outcomes \cite{samek2017explainable}. With our goal of incorporating improved semantics in the image classification task, we take inspiration from the past work on content-based image retrieval (CBIR) \cite{Worring2000, LIU2007262, MULLER20041}, as this addresses a few key points in general image understanding architectures. The image retrieval task aims to output accurate images given a user query. Zhou et al. \cite{ZhouLT2017} point out a crucial challenge in image retrieval as the \textit{semantic gap}, that deals with the effectiveness of translating semantic concepts to low-level visual features \cite{ZhouLT2017}. Addressing this \textit{semantic gap} in image understanding approaches is at the core of our study. Ying Liu et al. \cite{LIU2007262} presents a comprehensive study on the many approaches taken in bridging the \textit{semantic gap}, which includes machine learning as one of the major avenues. 

\begin{figure}[]
\centering
\includegraphics[scale=0.4]{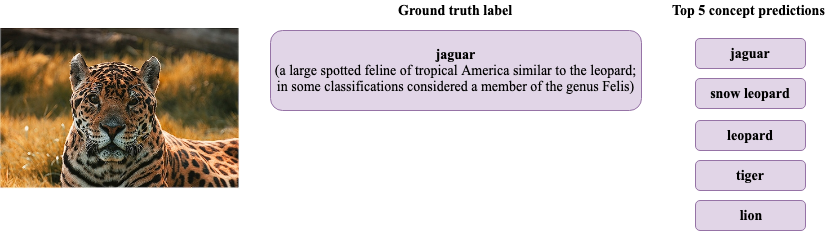}
\caption{Top 5 classification results of our system given an image of class \textit{jaguar}.}
\label{fig:fig0}
\end{figure}

In this study we employ a deep learning algorithm \cite{krizhevsky2012imagenet} for visual feature extraction purposes. Our approach to address the \textit{semantic gap}, makes use of concept information extracted from structured knowledge bases combined with supervised deep learning techniques. We adopt an image classification setting for in-depth evaluation of the approach. Figure \ref{fig:fig0} shows a snapshot of our proposed system's outcome, predicting the five closest matches of concepts to the given input image. While the image being accurately classified, we see a semantic similarity between all the other predicted concepts, which the system has learnt through the external knowledge used during training. 

Andrea Frome et al. \cite{frome2013devise} presents a method to incorporate word embeddings \cite{mikolov2013efficient} in a visual-semantic embedding model that is employed for image classification. They showcased the technique using 1000 classes of the ILSVRC 2012 dataset \cite{deng2009imagenet}, where the ground truth labels were replaced with word embeddings obtained from a skip-gram architecture \cite{mikolov2013efficient, mikolov2013distributed} while training. The results showed that the model could match state-of-the-art classification performance while making semantically reasonable errors. Most notably, it showed capabilities of zero-shot learning \cite{palatucci2009zero}. Several similar approaches have been presented since, that proposed methods of integrating word embeddings with vision architectures and improved performance in zero-shot learning \cite{zhang2017learning, zhang2015zero, norouzi2013zero}. Our study was formulated primarily inspired by the above contributions and with motivations towards improving the semantic representations employed. We aim to explore the effect of structured knowledge representations, such as knowledge graphs, in place of unstructured text corpora \cite{frome2013devise}, hoping to capture more explicit relationships between entities. We study the effects of such an approach in an image classification task. 

The main contribution of this study is the introduction of a technique to embed semantics learnt from structured knowledge bases in a visual-embedding model. We also introduce a novel evaluation framework to asses the generalisation of concepts with respect to the structured knowledge used to inform the semantics during training. We demonstrate zero-shot learning and introduce of a novel evaluation metric to assess its performance according to the class hierarchy of the structured knowledge used.

\section{Proposed method}

Our approach to incorporate semantics of structured knowledge bases into the image classification task can be subdivided into two main steps. First is a technique to obtain accurate concept representations, which will define the vector space for our architecture. Next is mapping visual features of images obtained via a visual-semantic embedding model to the vector space of the concept representations. The goal of this is to have both image representations and concept representations in a single vector space, so that they can be compared during the image classification task. 

\subsection{Obtaining concept representations}
\label{subsec:approach-concept}

\begin{figure}[h!]
\centering
\includegraphics[scale=0.5]{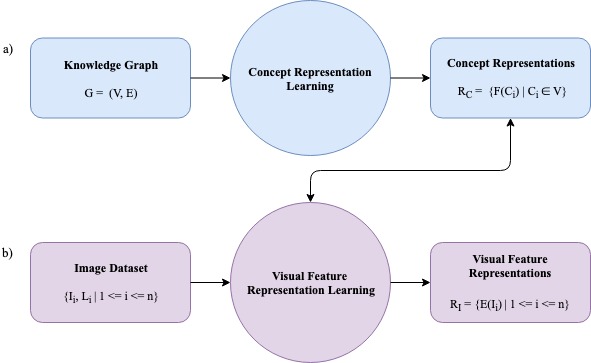}
\caption{Flow diagram showing the main two steps in our approach. a) Represents the process of obtaining concept representations from the knowledge base. b) Represents the process of mapping visual features to the obtained concept representations}
\label{fig:fig1}
\end{figure}

Concept representations define the vector space for our architecture. Figure \ref{fig:fig1} a) shows this process in a flow chart where, initially we have a graph $G$ with nodes $V$ and edges $E$. We assume either that all edges are is-a (directed specialisation) edges or some edges are labelled as such. The reason for this is that, we identify is-a as the basic form of a specialisation hierarchy that would be used to inform the semantics of the concepts in concern. Of course, moving forward the edges can have more complex relational representations, which is beyond the scope of our current study. Also we do not restrict the structure of the graph. An important thing to consider when selecting a graph is that, the nodes should contain the training image label classes of the classification task ($L_i$ in Figure \ref{fig:fig1} b)). And the graph should represent the relations of concepts to each other in the considered domain. 


Chakaveh Saedi et al. \cite{saedi2018wordnet} recently proposed a technique to compute embeddings, specifically on WordNet \cite{miller1998wordnet}, that could represent lexical semantics of the nodes. Their study showed how the resulting embedding vectors are superior to the performance of word embeddings \cite{mikolov2013efficient} in semantic similarity tasks. In our approach we make use of these findings to obtain the embeddings as our concept representations. 

$G$ is converted to an adjacency matrix $M$ such that, if two words in $G$, $w_i$ and $w_j$, are directly related by an edge, the entry $M_ij$ is set to 1 (otherwise 0). Also to account for words that are not directly connected to each other, $M$ is further enriched taking distantly connected nodes and aggregated as in Equation \ref{eq:eq1}.

\begin{equation}
\label{eq:eq1} 
    M_G = \sum_{n=0}^{\infty} (\alpha M)^{n} = (I - \alpha M)^{-1}
\end{equation}

Where, $n$ is the length of the path between two nodes, $\alpha$ ($<$ 1) is a decay factor that determines the effect of path length on $M$. Longer the path between two words (larger the $n$), lesser it affects $M_G$. $M_G$ is normalised using L2-norm and reduced to a set of vectors with a lower dimensionality using Principal Component Analysis (PCA). These vectors will be our concept representations. 

\subsection{Mapping visual features to concept representations}
\label{subsec:approach-mapping}

After defining the concept vector space as shown in Section \ref{subsec:approach-concept}, we want to train our visual-semantic embedding model to map input images to the relevant concept representations. Figure \ref{fig:fig1} b) summarises the procedure where, in the visual feature representation learning step, the learnt concept representations $R_C$ are used to inform the semantics of the concepts.

We employ a deep neural network here, with an additional project layer at the top of the network that outputs a representation matching the dimensionality of the concept representations. The projection layer is trained using the cosine similarity between the predicted embeddings and the pre-computed concept representations as a measure of loss according to Equation \ref{eq:eq2}.

\begin{equation}
\label{eq:eq2}
    loss(I,R_C) = 1 - \cos(E(I), R_C)
\end{equation}

In Equation \ref{eq:eq2}, $I$ and $R_C$ are the image and the corresponding concept representation of the image label class respectively. $E$ is the function that represents both feature extraction and mapping of the projection layer. We minimise the value of 1 minus the cosine similarity during training since the more similar the two vectors are, they should result in a cosine similarity closer to 1. After training, we use the function $E$ to calculate the visual embeddings of all the training images and store them. $R_I$ in Figure \ref{fig:fig1} b) represents the set of these embeddings.

\subsection{The evaluation frameworks}
\label{subsec:evaluationf}


\subsubsection{Taxonomy-Aware Measure for Errors (TAME)}
\label{subsec:subsumptiontest}
We introduce a novel evaluation framework which we name as \textit{Taxonomy-Aware Measure for Errors (TAME)}, that analyses the classification errors with respect to the property of $subsumption$ extracted from the hierarchical structure of the knowledge base used. $Subsumption$ is formed by the is-a relations in the knowledge base, for example, if $C$ is related to $D$ via an is-a link, we can say that $D$ $subsumes$ $C$ and that all properties of $C$ is also present in $D$. We make use of this property to check if our architecture is effective in generalising to the concepts that subsumes the ground-truth classes of the images. We can check this at different levels of subsumption going up the is-a hierarchy.



\begin{figure}[h!]
\centering
\includegraphics[width=0.70\linewidth]{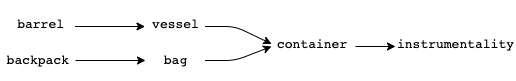}
\caption{The hypernym hierarchy of the classes $barrel$ and $backpack$ up to 3 subsumption levels}
\label{fig:fig6}
\end{figure}

In the presence of background knowledge, it makes sense to distinguish \textit{bad} from \textit{not so bad} errors; for example, if the system predicts a subsumer class of the ground-truth class instead of the ground-truth class itself, this is a \textit{less bad} error than if it predicted a less closely related class. For example, referring to Figure \ref{fig:fig6}, if the system classifies an image of a \textit{barrel} as a \textit{vessel}, TAME will consider it to be \textit{almost correct}. TAME captures this aspect of our system and quantifies image classification results with semantically accurate errors. Results are shown in Section \ref{subsec:subsumption}.

\subsubsection{Addition to zero-shot evaluation}
\label{subsec:zero-shotextension}

Next a novel approach to evaluate zero shot learning is presented. We divide the set of chosen zero-shot classes into two sets, namely, $Sibling$ and $NonSibling$ classes. The notion of being $sibling$ classes is derived from the property of sharing common subsumer classes. For example, referring to Figure \ref{fig:fig6}, $vessel$ and $bag$ share the same subsumer, $container$. Hence we categorise such classes as $Sibling$ classes in the zero-shot evaluation, while the other being $Nonsibling$ classes. During evaluation, we distinguish the difference in performance of these two sets in the image classification task. 



\section{Experiments}
\label{sec:EEvaluation}

\begin{figure*}[h]
\centering
\includegraphics[width=0.8\linewidth]{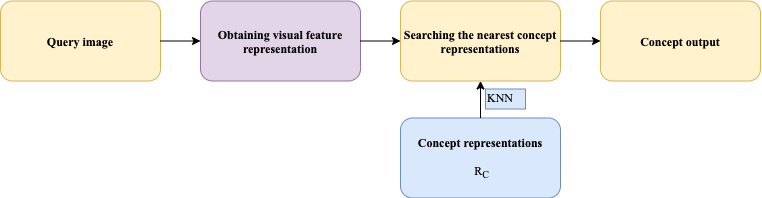}
\caption{System setup to retrieve a concept given an image as query (Image classification). $R_C$ is the same vector as in Figure \ref{fig:fig1}. }
\label{fig:fig2}
\end{figure*}

In this study, we make use of ILSVRC 2012  \cite{russakovsky2015imagenet} as the image dataset and WordNet \cite{miller1998wordnet} as the knowledge base for obtaining concept representations. Since ILSVRC 2012 uses WordNet entities as the ground truth class labels for the objects in the images, it qualifies as a reliable structured knowledge base for all the concepts found in ILSVRC 2012. WordNet is a lexical ontology for English language that consists over 120k concepts, 25 types of relations between these concepts and over 155k words (lemmas) that are categorised to nouns, verbs, adjectives and adverbs. In this implementation, we adopt the same approach of \cite{saedi2018wordnet} and extract a sub graph of 60k words from all parts-of-speech in WordNet. All relations were considered and weighted equally during the embedding calculation according to Section \ref{subsec:approach-concept}. The dimensionality is chosen to be 850 for the resulting concept representations \cite{saedi2018wordnet}. We employ a deep residual network \cite{he2016deep} (Resnet 50), pre-trained for an image classification task with the ILSVRC 2012 dataset. The softmax prediction layer at the top of the network is replaced with a projection layer, as explained Section \ref{subsec:approach-mapping}. The projection layer acts as a linear transformation that maps a 512 dimensional feature vector of the image to a 850 dimensional concept embedding produced in our previous step. To evaluate our goal of semantically informed image classification, we use the process shown in Figure \ref{fig:fig2}. 

Throughout our evaluation we compare the results with the DeViSE \cite{frome2013devise} architecture trained on the same classes and tested against the same settings. DeViSE takes a similar approach to embed semantics in image classification with the only difference of concept representations being word embeddings \cite{mikolov2013efficient}. This gives us the opportunity to compare the difference in performance when concept representations are calculated using structured knowledge versus unstructured text.  

\subsection{Evaluating visual model performance}
During the training phase of visual-semantic embedding model (denoted by $E$ in Figure \ref{fig:fig1} b)), we choose 300 randomly selected classes out of the ILSVRC 2012 1K dataset \cite{russakovsky2015imagenet}. The reason behind selecting the 300 classes was that, the nodes of WordNet graph considered during the concept extraction phase included these 300 classes. We compare the Hit@$k$ image classification accuracies of our approach with a standard Softmax image classification network baseline (Resnet50 \cite{he2016deep}) and DeViSE \cite{frome2013devise}, both trained on the same training classes. The results are shown in Table \ref{tab:tab1}.

\begin{table}
{\caption{Hit@$k$ accuracies of our system compared to Softmax classifier and DeViSE}\label{tab:tab1}}
\centering
\begin{tabular}{lccc}
\hline
\multirow{2}{*}{\textbf{Model}} & \multirow{2}{*}{\textbf{Dataset}} & \multicolumn{2}{c}{\textbf{Hit@k(\%)}} \\ \cline{3-4} 
                                &                                   & \textbf{1}         & \textbf{5}        \\ \hline
\multirow{2}{*}{Softmax}        & Train Set                         & \textbf{89.06}              & \textbf{94.14}             \\ \cline{2-4} 
                                & Test Set                          & 46.69              & 63.55             \\ \hline
\multirow{2}{*}{DeViSE}         & Train Set                         & 60.67              & 70.88             \\ \cline{2-4} 
                                & Test Set                          & 53.36              & 66.20             \\ \hline
\multirow{2}{*}{Proposed}       & Train Set                         & 71.92              & 88.93             \\ \cline{2-4} 
                                & Test Set                          & \textbf{56.82}              & \textbf{77.06}             \\ \hline
\end{tabular}
\end{table}

From the results in Table \ref{tab:tab1}, we see that although the softmax model attains higher accuracies in image classification for the training set, both DeViSE and our proposed model show higher accuracies for the test set, with our model having the highest accuracy in both cases of Hit@1 and Hit@5. The test set consists of unseen images of the same training classes and higher accuracies with this set implies the superior generalisation ability of the systems. Nevertheless, for the focus on this study we understand that our visual-semantic embedding model, in fact has promising performance in image classification. 

In addition to the above results, our system inherits the capability of incorporating semantics of the classes with their relationships according to the structured knowledge base used during training (i.e. WordNet). We move on to our novel evaluation framework introduced in Section \ref{subsec:subsumptiontest} to evaluate this capability.

\subsection{TAME on image classification}
\label{subsec:subsumption}

ILSVRC 2012 provides the hierarchy of the selected classes according to the hypernym tree in WordNet, similar to the example is shown in Figure \ref{fig:fig6}. We obtain three sets of subsumers for all the 300 classes selected, at three levels, each above the 300 training classes, named as 1-step, 2-step and 3-step subsumers. As the levels increase, the classes become more general in meaning. Note that these subsumer classes are not used during the training of the systems. The system will have to deduce the semantic relationship informed by the concept representations provided during the training process in order to accurately identify the more general subsumer classes. The prediction is taken as correct if the system outputs either the ground truth class or any subsumer class in each of the tests. The results are compared the DeViSE \cite{frome2013devise} retrained using the same 300 training classes and tested with the same subsumer sets. The results are shown in Table \ref{tab:table3}.


\begin{table*}[th]
 {\caption{Image classification performance including subsumer classes. 1-step, 2-step and 3-step subsumers represent classes in respective number of steps above the ground truth class labels according to the WordNet hypernym (is-a) hierarchy.}\label{tab:table3}}
 \centering
 \resizebox{\textwidth}{!}{%
\begin{tabular}{llccccccccccccccc}
\hline
\multicolumn{1}{c}{\textbf{Model}}    & \multicolumn{8}{c}{\textbf{DeViSE}}                                                                       & \multicolumn{8}{c}{\textbf{Proposed}}                                                                         \\ \hline
\multicolumn{1}{c}{\textbf{Dataset}}  & \multicolumn{4}{c}{\textbf{Train Set}}              & \multicolumn{4}{c}{\textbf{Test Set}}               & \multicolumn{4}{c}{\textbf{Train Set}}              & \multicolumn{4}{c}{\textbf{Test Set}}               \\ \hline
\multicolumn{1}{c}{\textbf{Accuracy}} & \multicolumn{4}{c}{\textbf{Hit@k\%}}                & \multicolumn{4}{c}{\textbf{Hit@k\%}}                & \multicolumn{4}{c}{\textbf{Hit@k\%}}                & \multicolumn{4}{c}{\textbf{Hit@k\%}}                \\ \hline
\textbf{}                             & \textbf{1} & \textbf{5} & \textbf{10} & \textbf{20} & \textbf{1} & \textbf{5} & \textbf{10} & \textbf{20} & \textbf{1} & \textbf{5} & \textbf{10} & \textbf{20} & \textbf{1} & \textbf{5} & \textbf{10} & \textbf{20} \\ \hline
1-step Subsumers                      & 60.94      & 75.75      & 80.55       & 84.65       & 53.36      & 66.20      & 68.45       & 70.70       & 72.24      & 88.79      & 91.03       & 95.02       & 61.29      & 80.17      & 83.39       & 88.06       \\ \hline
2-step Subsumers                      & 60.87      & 76.92      & 82.68       & 87.25       & 53.41      & 72.70      & 79.57       & 84.70       & 72.26      & 88.79      & 91.12       & 95.13       & 61.41      & 80.20      & 83.63       & 88.39       \\ \hline
3-step Subsumers                      & 60.77      & 76.81      & 82.88       & 87.86       & 55.94      & 73.58      & 79.86       & 85.09       & 73.65      & 90.24      & 92.30       & 95.99       & 62.61      & 81.72      & 84.98       & 89.40       \\ \hline
\end{tabular}%
}

\end{table*}

The results demonstrate that both architectures have the ability to generalise to subsumer classes from the trained classes. This is seen from the increasing accuracy along with the inclusion of 1-step, 2-step and 3-step subsumers of both architectures. It can also be understood that both the approaches gain a semantic understanding about the classes that is not captured during the traditional classification test (results in Table \ref{tab:tab1}). The superior performance of our approach in almost all cases of TAME compared to DeViSE, shows the effectiveness of employing a structured knowledge rather than unstructured text in obtaining the semantics used to train the visual-semantic embedding model. Although embeddings learned from unstructured text also attains a level of generalisation to similar concepts, structured knowledge is able to capture these semantics more accurately.

\subsection{Evaluating zero-shot learning}
\label{subsec:zeroshot}

To demonstrate zero shot learning, we pick a random set of 30 totally unseen classes from ILSVRC 2012 dataset. We identify the sibling properties of these classes, as explained in Section \ref{subsec:zero-shotextension}, with respect to the 300 classes used during training. It turned out that 14 out of the 30 zero-shot classes were sibling classes and the rest were Non-sibling classes. 

We use the same DeViSE architecture from Section \ref{subsec:subsumption} for comparison of results. The tests were carried out under two settings of $R_C$ (from Figure \ref{fig:fig2}), where results were taken with and without $R_C$ containing the training class embeddings. The classification is considered as correct if the correct label is present among the $k$ outputs in each Hit@$k$ task. The results are shown in Table \ref{tab:table4}.

\begin{table*}[th]
{\caption{Zero-shot learning results for both sibling and non-sibling classes}\label{tab:table4}}
\centering
\begin{tabular}{lcccccc}
\hline
\multirow{3}{*}{\textbf{Model}}              & \multicolumn{3}{c}{\textbf{Sibling Classes}} & \multicolumn{3}{c}{\textbf{Non-Sibling classes}} \\ \cline{2-7} 
                                             & \multicolumn{3}{c}{\textbf{Hit@k(\%)}}       & \multicolumn{3}{c}{\textbf{Hit@k(\%)}}           \\ \cline{2-7} 
                                             & \textbf{1}    & \textbf{5}   & \textbf{10}   & \textbf{1}     & \textbf{5}     & \textbf{10}    \\ \hline
DeViSE (Only zero-shot class labels)         & 48.63         & 60.66        & 71.86         & \textbf{28.07}          & 48.40          & 69.52          \\ \hline
DeViSE (Zero shot + training class labels)   & 0.55          & 14.48        & 28.96         & \textbf{0.27}           & 6.42           & \textbf{16.31}          \\ \hline
Proposed (Only zero-shot class labels)       & \textbf{53.01}         & \textbf{74.04}        & \textbf{89.07}         & 25.67          & \textbf{54.28}          & \textbf{77.27}          \\ \hline
Proposed (Zero shot + training class labels) & \textbf{1.64}          & \textbf{27.60}        & \textbf{41.53}         & 0.00           & \textbf{8.82}           & 12.83          \\ \hline
\end{tabular}
\end{table*}

Both models demonstrate capabilities of zero-shot learning, with higher accuracies when $R_C$ does not contain the training class embeddings. The reason for this is that, the models map the zero-shot image embeddings more towards the known class labels when they are present in the system. One takeaway from the results in Table \ref{tab:table4} is that, the sibling classes are more accurately classified in both models when compared with non-sibling classes. The intuition behind this being the model's awareness of similar classes during training to the zero-shot classes (effect of having common subsumers). Our proposed model outperforms DeViSE in all instances in the sibling category. 
Another interesting insight is how our proposed model performs worse with non-sibling classes in several instances compared to DeViSE. This can be seen as a result of the structure of knowledge enforced on our embeddings. Next we extended the zero-shot evaluation to include TAME introduced in Section \ref{subsec:subsumptiontest}. 

\subsection{TAME on zero-shot evaluation}
\label{subsec:zeroandsubsumtion}

We extend our zero-shot evaluation to test if the models can generalise zero-shot classification also to the subsumer classes of the zero-shot classes. The subsumer classes are obtained by the same method as explain in Section \ref{subsec:subsumption} for the training classes. 
The results are shown in Table \ref{tab:table5}.

\begin{table}[ht]
{\caption{Results after extending the zero-shot results to include TAME.}\label{tab:table5}}
\resizebox{\textwidth}{!}{%
\begin{tabular}{ccccccccccccc}
\hline
\textbf{Model}            & \multicolumn{12}{c}{\textbf{DeViSE}}                                                                                                                                                                                    \\ \hline
\textbf{Dataset}          & \multicolumn{3}{c}{\textbf{Sibling Class}}   & \multicolumn{3}{c}{\textbf{Sibling + Training Classes}} & \multicolumn{3}{c}{\textbf{Non-Sibling Classes}} & \multicolumn{3}{c}{\textbf{Non-Sibling + Training Classes}} \\ \hline
\textbf{Accuracy}         & \multicolumn{3}{c}{Hit@k(\%)}                & \multicolumn{3}{c}{Hit@k(\%)}                           & \multicolumn{3}{c}{Hit@k(\%)}                    & \multicolumn{3}{c}{Hit@k(\%)}                               \\ \hline
\textbf{}                 & 1             & 5             & 10           & 1                & 5                 & 10               & 1              & 5              & 10             & 1                 & 5                  & 10                 \\ \hline
\textbf{1-step Subsumers} & 61.48         & 79.51         & 82.24        & 1.09             & 28.69             & 45.08            & 44.92          & 77.54          & 86.90          & 0.27              & 13.10              & 38.50              \\ \hline
\textbf{2-step Subsumers} & 50.55         & 77.05         & 81.42        & 1.09             & 30.60             & 50.55            & 37.17          & 77.54          & 88.50          & 0.27              & 17.38              & 42.25              \\ \hline
\textbf{3-step Subsumers} & 50.27         & 84.70         & 96.45        & 1.09             & 36.34             & 56.56            & 36.10          & 77.01          & 88.50          & 0.27              & 18.45              & 41.71              \\ \hline
\multicolumn{13}{c}{}                                                                                                                                                                                                                               \\ \hline
\textbf{Model}            & \multicolumn{12}{c}{\textbf{Proposed}}                                                                                                                                                                                      \\ \hline
\textbf{Dataset}          & \multicolumn{3}{c}{\textbf{Sibling Classes}} & \multicolumn{3}{c}{\textbf{Sibling + Training Classes}} & \multicolumn{3}{c}{\textbf{Non-Sibling Classes}} & \multicolumn{3}{c}{\textbf{Non-Sibling + Training Classes}} \\ \hline
\textbf{Accuracy}         & \multicolumn{3}{c}{Hit@k(\%)}                & \multicolumn{3}{c}{Hit@k(\%)}                           & \multicolumn{3}{c}{Hit@k(\%)}                    & \multicolumn{3}{c}{Hit@k(\%)}                               \\ \hline
\textbf{}                 & 1             & 5             & 10           & 1                & 5                 & 10               & 1              & 5              & 10             & 1                 & 5                  & 10                 \\ \hline
\textbf{1-step Subsumers} & 52.46         & 73.77         & 84.70        & 2.73             & 28.69             & 43.17            & 24.87          & 46.26          & 65.78          & 0.00              & 9.36               & 15.24              \\ \hline
\textbf{2-step Subsumers} & 53.55         & 73.22         & 83.61        & 3.01             & 28.96             & 44.26            & 26.74          & 51.60          & 65.51          & 0.00              & 13.37              & 20.86              \\ \hline
\textbf{3-step Subsumers} & 51.71         & 70.86         & 82.29        & 3.14             & 27.43             & 42.29            & 26.47          & 52.41          & 66.04          & 0.00              & 13.37              & 21.39              \\ \hline
\end{tabular}%
}
\end{table}

We take the same subsets of the Sibling and the Non-Sibling classes of the zero-shot set in Table \ref{tab:table4} and include 1-step, 2-step and 3-step subsumers of those classes to obtain the Hit@k classification accuracy. This evaluates the generalisation abilities of our architecture on the totally unseen zero-shot classes. With the results presented in Table \ref{tab:table5}, again we first see the better overall performance of the sibling classes of both models, going in-line with the results of Table \ref{tab:table4}. An observation to note is the decrease in accuracies seen with both approaches in several Hit@k instances with increasing step. The reason for this is that, as the subsumer steps increase, more concept representations are introduced to $R_C$ for the systems to choose from. Hence it increases the possibility of the models choosing wrong concepts for the zero-shot classes. 


\section{Related work}

Image understanding is often challenged by the accurate semantic understanding of concepts \cite{ZhouLT2017, MULLER20041} according to human-level knowledge. As Wengang Zhou et al. \cite{ZhouLT2017} points out, translation of human-level semantic knowledge to low-level visual feature representations is a crucial hurdle to overcome which is referred as the $semantic$ $gap$ \cite{ZhouLT2017}. Even though techniques of machine learning and deep learning have been applied to solving this problem \cite{LIU2007262}, accurate definition of human-level knowledge is lacking in the existing systems. 


For the goal of bridging the $semantic$ $gap$ we turn to the many approaches presented in the area of Zero-shot learning \cite{palatucci2009zero}. These are known for incorporating attributes to represent concepts in vector spaces and mapping visual features to them, giving more meaning to features extracted from an image \cite{farhadi2009describing, lampert2009learning}. Many recent deep learning inspired techniques of zero-shot learning make use of semantic embeddings learnt from a language base in an unsupervised setting, which are then used to guide visual feature representations \cite{socher2013zero, romera2015embarrassingly, fu2016semi, zhang2015zero, zhang2016zero}. Our study was largely inspired by the findings of Andrea Frome et al. \cite{frome2013devise}, where word embeddings learnt from unstructured text \cite{mikolov2013efficient} were employed for zero-shot classification of ILSVRC 2012 1K dataset \cite{russakovsky2015imagenet}. Our approach differs when we replace the word embeddings with concept representations captured via a structure knowledge base (such as a knowledge graph) to inform the visual-semantic embedding model.


Xiaolong Wang et al. \cite{wang2018zero} recently explored the idea of incorporating relations from knowledge graphs into zero-shot learning with the use of Graph Convolutions Networks (GCN) \cite{kipf2016semi}. They used the relation information from the edges in the graph to enrich the word embeddings in the process of inferring visual features of unseen images. Our study differs from this approach, where we completely replace the word embeddings with embeddings produced from a structured knowledge base that encompasses the relations between the entities concerned.

\section{Conclusion}

We propose a technique to incorporate concept representations obtained from a structured knowledge base to inform a visual-semantic embedding model. We show how these embeddings are calculated using the graph structure of the knowledge base and also how they are used to train a projection layer at the top of a traditional image classification model. Informed by the semantics, the image classification model is able to attain higher generalisation with enhance classification accuracies in the test sets compared with traditional methods. We introduce two novel evaluation frameworks that demonstrates how to assess the classification errors with respect to the structure of the knowledge base used. Our approach shows promising results in TAME, where the ability of our approach to identify more general classes of the ground-truth classes was evaluated according to the class hierarchy.

We demonstrate zero-shot classification and evaluate its performance with the novel addition of notions of sibling and non-sibling classes. The results show how sibling classes perform better in a zero-shot classification setting and how the structure of the external knowledge play a part in the classification.  

Overall with the results, we identify how concept representations extracted from structured knowledge are more effective than ones from unstructured text when informing a visual-semantic embedding model.

\bibliographystyle{alpha} 

\bibliography{samplebib}

%
%
%

\end{document}